# Introduction to Camera Pose Estimation with Deep Learning

Yoli Shavit
Toga Networks a Huawei company
yoli.shavit@huawei.com

Ron Ferens
Toga Networks a Huawei company
ron.ferens@huawei.com

**Abstract**

*Over the last two decades, deep learning has transformed the field of computer vision. Deep convolutional networks were successfully applied to learn different vision tasks such as image classification, image segmentation, object detection and many more. By transferring the knowledge learned by deep models on large generic datasets, researchers were further able to create fine-tuned models for other more specific tasks. Recently this idea was applied for regressing the absolute camera pose from an RGB image. Although the resulting accuracy was sub-optimal, compared to classic feature-based solutions, this effort led to a surge of learning-based pose estimation methods. Here, we review deep learning approaches for camera pose estimation. We describe key methods in the field and identify trends aiming at improving the original deep pose regression solution. We further provide an extensive cross-comparison of existing learning-based pose estimators, together with practical notes on their execution for reproducibility purposes. Finally, we discuss emerging solutions and potential future research directions.*

1. Introduction

Recovering the camera-to-world translation and orientation from an image is one of the fundamental problems in computer vision. Accurately estimating the absolute pose of the camera is key to applications of augmented reality, autonomous navigation and robotics, where localization is crucial for performance. In recent years, deep Convolutional Neural Networks (dCNNs) have demonstrated an impressive success in learning different computer vision tasks such as image classification [1,2], object detection [3,4] and semantic segmentation [5,6].

Leveraging on the idea of transfer learning, pre-trained dCNNs were further used to learn other visual tasks that have a limited amount of training data but share the same low level features. Motivated by these advances, Kendall *et al.* proposed PoseNet [7], a modified truncated GoogLeNet [3] architecture where softmax classification is replaced with a sequence of fully connected layers, to output the absolute pose of a camera from an image. PoseNet was the first learning-based architecture that introduced the idea of regressing the absolute pose with a deep architecture. This approach offered several appealing advantages compared to classical structure-based methods: short inference times (milliseconds instead of minutes), low memory footprint (megabytes compared to gigabytes) and without any feature engineering required. In addition, by using transfer learning, the network is able to learn from limited-sized datasets.

However, PoseNet over-fitted its training data and could not generalize well to unseen scenes. Furthermore, its localization error on indoor and outdoor datasets was an order of magnitude larger, compared to feature-based approaches that are considered *state-of-the-art* [8,9]. These limitations motivated a surge of deep learning-based methods for absolute pose estimation (APE).

In this paper we provide a guided tour for visual pose estimation with deep learners. Our review is written for computer vision and deep learning researchers who are new to deep APE. We first lay the groundwork for our tour, defining the problem, the main evaluation metrics and the starting-point architecture (PoseNet). Next, we identify current trends, originating from different hypotheses on the modifications required to improve the original solution. Specifically, we propose a hierarchical grouping of deep learning solutions, based on their approach to the pose estimation problem, which facilitates their analysis. We describe representative architectures in each of the identified groups and discuss the assumptions that motivated their development. We further provide a comprehensive cross-comparison of over 20 algorithms, which includes a compilation of results (localization error) over several datasets as well as other characteristics of interest (e.g., reported runtime). We analyze the pros and cons of different solutions considering accuracy and other metrics that are relevant for real-life applications. In order to shorten time-to-implementation for applied researchers, we also include links to available open source implementations and provide key guidelines for when such implementation is not available. Finally, we discuss the current limitations and challenges of deep learning-based pose estimation and suggest promising directions for future

work. In summary, our contributions are as follows:
- A guide to absolute pose estimation with deep learning, providing both theoretical background and practical advice.
- Cross-comparison of performance and characteristics of over 20 deep learning pose estimators.
- Summary of existing and emerging trends in deep pose estimation, and the current challenges and limitations.

### 1.1. Problem Definition

Given an image $I_c$, captured by a camera $C$, an absolute pose estimator $E$ tries to predict the 3D pose orientation and location of $C$ in world coordinates, defined for some arbitrary reference 3D model (a 'scene').

The translation of $C$ with respect to the origin (location) is specified by a vector $x \epsilon \mathbb{R}^3$. The orientation of $C$ can be described with several alternative representations, such as a $3x3$ rotation matrix, quaternion and Euler angles. Most commonly, the quaternion representation is used, specifying the orientation as a vector $q \epsilon \mathbb{R}^4$. This representation elevates the need for orthonormalization, which is required for rotation matrices, and can be converted to a (legitimate) rotation when normalizing it to unit length [7]. One caveat of the quaternion representation is its potential ambiguity, due to a dual mapping of the quaternions q and –q to the same rotation operation. A variant of Euler angle has been used to address this problem in some solutins [10]. In practice, however, the majority of pose estimators predicts the quaternion representation (for a more extended review of the different representations for pose orientation, see [11]). The overall pose of $C$ is thus specified with a tuple $p = (x, q)$.

The APE problem can now be formally defined as the problem of estimating a function $E$ taking an image $I_c$ captured by a camera $C$ and outputting its respective pose:

$$E(I_c) = (x_c, q_c) \quad (1)$$

Note that the definition given in Eq. 1 can be extended to additional inputs about the camera and the image (e.g., depth and camera frustum).

A related problem, which is often solved jointly or in parallel to APE (for example in visual odometry systems), is the relative pose estimation (RPE) problem. In a RPE setting, the estimator takes two images, $I_c^1$ and $I_c^2$, captured by $C$ and aims to predict the relative pose between them. Eq. (1) can be modified to capture this problem:

$$E(I_c^1, I_c^2) = (x_c^{rel}, q_c^{rel}) \quad (2)$$

### 1.2. Evaluation Metrics

In order to evaluate the performance of a pose estimator, we require a set of images and the ground truth poses of the camera(s) which captured them. Since the camera pose is related to some 3D model coordinates, such a model needs to be available. Typically a 3D point cloud, associated with a set of images for training and testing, is provided either through the scanning device (e.g., Microsoft Kinect) or through reconstruction using structure-from-motion (SfM) methods. Popular SfM tools are Bundler [12], COLMAP [13,14] and VisualSFM [15].

Given a ground truth pose $p = (x, q)$ and an estimated pose $\hat{p} = (\hat{x}, \hat{q})$, the localization error of $\hat{p}$ is measured by the deviations between the translation (location) and rotation (orientation) of $p$ and $\hat{p}$.

The translation error $t_{err}$ is typically measured in meters and defined as the Euclidian distance between the ground truth and estimated locations:

$$t_{err} = \Delta x = \|x - \hat{x}\|_2 \quad (3)$$

The rotation error $rot_{err}$ is typically measured in degrees and corresponds to the minimum rotation angle α required to align the ground truth and estimated orientations [16,17]:

$$2\cos(\alpha) = \text{trace}(R^{-1}\hat{R}) \quad (4)$$

Where $R$ and $\hat{R}$ are the ground truth and estimated $3x3$ rotation matrices, respectively, and $tr(M)$ is the trace of $M$. Using the quaternion representation, $rot_{err}$ is given by:

$$rot_{err} = \alpha = 2\cos^{-1}|q\hat{q}|\frac{180}{\pi} \quad (5)$$

The relative pose error is computed in a similar manner to the absolute pose error, based on the deviation between the ground truth and estimated relative poses. It is typically measured in [*m/s*] and [*degree/s*] (for translation and rotation, respectively), capturing the drift when computed over a sequence.

The translation and rotation errors are commonly reported as a summary statistics (e.g., the median). Alternatively, some papers report the localization rate, defined by computing the percentage of images localized within a given translation and rotation error thresholds (for example, with translation and rotation errors smaller or equal to 0.25 meters and 2 degrees).

### 2. Deep Architectures for visual absolute pose estimation

Traditionally, visual APE has been achieved with image retrieval or structure-based approaches. Structure-based methods typically rely on SfM (hence the name) to localize. Specifically, SfM associates 3D points with 2D images that capture them and with their local descriptors (found through image processing). Matches between 2D points in the image and 3D points in the scene are then found by searching through the shared descriptor space. Note that the descriptors can either be hand crafted (e.g., SIFT [18]) or learned (e.g., SuperPoint [19]). Given 2D-3D matches, a n-point-pose (PnP) solver estimates candidate poses, and the best pose hypothesis is chosen using RANSAC [20]. The estimated best pose is typically subjected to a further refinement.

In image retrieval, a query image (for which a pose should be estimated) is used to search against a database of images with known ground truth poses. The pose of the query image is taken to be the pose of the nearest neighbor. Similarly to local descriptors (used in structure-based methods), global descriptors may also be feature-based or learned. *State-of-the-art* performance, in terms of speed and accurate retrieval, has been achieved with deep architectures [21,22], generating efficient image encodings. However, in the context of absolute pose estimation, even *state-of-the-art* retrieval methods provide only a rough estimate of the ground truth pose and thus mainly serve for place recognition [23], rough localization [24] or as a baseline for APE evaluation [25].

Instead of learning a global image descriptor for retrieval or matching local 2D-3D features for PnP-based pose estimation, Kendall *et al.* suggested to learn the localization pipeline in its entirety. Starting from an RGB image, a network can learn to regress the camera pose in an end-to-end supervised manner (i.e., given the ground truth pose). Under the assumption that low level features used for general vision tasks (e.g., image classification) encode useful information for pose estimation, the researchers suggested to leverage on transfer learning from pre-trained dCNNs for pose estimation. The suggested architecture, named PoseNet, is a modification of a GooLeNet architecture, a deep convolutional network with 22 layers (six Inception modules), where softmax layers (for classification) were replaced with deep fully connected (FC) layers to regress the pose. Specifically, in PoseNet, the GooLeNet network is truncated after the average pooling layer, outputting a 1024-dimensional vector $v$, $v \epsilon \mathbb{R}^{1024}$, which represents an encoding of the visual features of the input image. The vector $v$ is then fed to a 2048-dimensional FC layer that maps it to a localization feature vector, $u \epsilon \mathbb{R}^{2048}$. Finally, the translation and quaternion vectors $\hat{x}$ and $\hat{q}$ are regressed from $u$ with two separate FC layers, to give the estimated pose $\hat{p} = (\hat{x}, \hat{q})$. The described pose regression architecture can be abstracted into three components: an encoder which generates the visual encoding vector $v$, a localizer which outputs the localization features vector $u$ and a regressor which regresses pose $\hat{p}$. Fig. 1 shows a schematization of the main building blocks (encoder, localizer and regressor) of the PoseNet's architecture.

PoseNet was trained with Stochastic Gradient Descent (SGD) to minimize the following *pose loss* function:

$$\mathcal{L}_\beta(I_c) = \mathcal{L}_x(I_c) + \beta \mathcal{L}_q(I_c) = \|x - \hat{x}\|_2 + \beta \|q - \frac{\hat{q}}{\|\hat{q}\|_2}\|_2 \quad (6)$$

where $\mathcal{L}_x$ gives the translation loss (identical to the translation error, see Eq. 3), $\mathcal{L}_q$ gives the rotational loss for $\beta$ is a scaling term balancing between the two losses. Note that the set of rotation matrices map to quaternions of unit length (a unit length sphere). This implies that for $\hat{q}$ to map to a legitimate rotation it must be normalized to unit length (i.e., $\hat{q} = \frac{\hat{q}}{\|\hat{q}\|_2}$). In practice, PoseNet's authors removed this constraint from the optimization in their original implementation after observing that $\hat{q}$ came close enough to q, even without normalization (de-facto satisfying the sphere constraint)[7]. To maintain validity, at test time, the estimated quaternion was normalized to unit length.

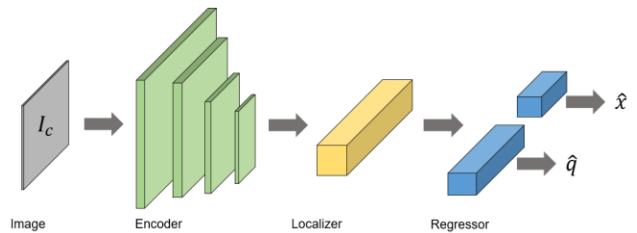

**Figure 1.** A schematization of the PoseNet's architecture. Given an image $I_c$, a dCNN architecture ('Encoder') generates visual feature vectors from $I_c$. Using a FC layer ('Localizer'), the visual encoding of $I_c$ is mapped to a localization feature vector. Finally, two separate connected layers ('Regressor') are used to regress $\hat{x}$ and $\hat{q}$, respectively, giving the estimated pose $\hat{p} = (\hat{x}, \hat{q})$. A similar abstraction was previously suggested by Sattler *et al.* [25].

The idea of regressing the absolute pose with end-to-end learning offered several appealing advantages compared to traditional structure-based methods. Deep absolute pose regression does not require any feature engineering and rely on (dCNN based) encodings that were shown to be more robust to challenging changes in the scene, such as lighting conditions and viewpoint [26]. In comparison with structure-based methods, which require a 3D model and heavy computations online (2D-3D matching and PnP inside a RANSAC loop), a trained model has a low memory footprint and constant runtime at inference. In addition, transfer learning enabled effective training on commonly used medium-sized datasets [27,7]. However, the localization error (translation and rotation) achieved with PoseNet was an order of magnitude larger than the error attained with *state-of-the-art* structure-based methods (see Tables 3-4). In addition, issues such as generalization to unseen scenes and the learning capacity of the model were not fully addressed in the original work.

## 2.1. Beyond Absolute Pose Regression

The limited ability of PoseNet to accurately estimate the absolute camera pose from an image led to the development of different deep learning-based APEs (deep APEs). In this review, we suggest that these methods could be grouped in a hierarchical manner (Fig. 2). At a high-level, deep learning estimators, take either an (1) end-to-end learning approach (Section 2.1.1) or a (2) hybrid approach (Section 2.1.2). Within the end-to-end learning cluster, we identify 2 main algorithmic groups: (1a) Pose regressors such as PoseNet and other modifications to its architecture and/or loss; (1b) Auxiliary learners, which jointly learns APE with auxiliary tasks, such as visual odometry and semantic segmentation. The hybrid cluster instead, includes methods that learn related problems, and combine them with other techniques to estimate the absolute pose. This is done by (2a) learning only local sub-tasks and coupling them with a structure-based localization pipeline; (2b) learning the relative camera pose between images in order to recover the absolute pose from nearest-neighbors (leveraging on the image retrieval paradigm); (2c) taking a hierarchical approach and combining image retrieval and structure-based learning-based methods. In the following sections we discuss each algorithmic cluster in more details and describe key methods within each cluster.

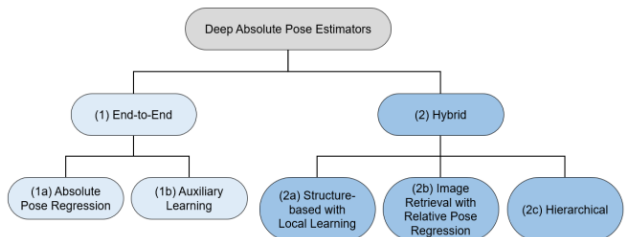

**Figure 2.** A hierarchical clustering of learning-based methods for APE. At the top level, current methods can be roughly grouped into (1) end-to-end and (2) hybrid learning approaches. Within each of the two clusters, more specific approaches have been developed (see Section 2.1.1 and Section 2.1.2 for more details). Tables 2-4 provide a cross comparison of representative algorithms for deep absolute pose estimation, considering pose error and other properties.

### 2.1.1 End-to-end learning of pose estimation

Shortly after the publication of PoseNet, Kendall and Cipolla leveraged on the notion of Bayesian CNNs with Bernoulli distributions [28] to create a Bayesian PoseNet [29]. Given an image and a pre-trained PoseNet model, a sample is generated by dropping out activation units (of convolutional layers) with a given probability. The pose is then computed by averaging over the individual samples' predictions. The covariance of these predictions further provides a measure for model uncertainty, which is correlated with pose error [29]. This change improved PoseNet's accuracy by 10% to 20%, on average (see Tables 3-4). Soon to follow, other pose regression methods, focused on modifications to the architecture and/or the loss function in order to improve PoseNet's original performance.

*Pose Regressors*

Rather than addressing overfitting by regularizing the dCNN model (Bayesian PoseNet), Walch *et al.*, suggested to address this problem by adding four Long-Short-Term-Memory (LSTM) layers after the 2048-dimensional FC layer (the Localizer component, see Section 2.1 and Fig. 1) [30]. In this architecture, each LSTM layer operates independently in one of four directions (right, left, up, down) in order to reduce the dimensionality of the image encoding (visual feature vector outputted by the dCNN encoder) and constrain the learning. The underlying assumption motivating this change was that the high-dimensionality of the image encoding, compared to the relatively small number of training examples, leads to overfitting since the final FC layers (regressor) need to learn a regression problem with many degrees of freedom. [30].

Several other modifications to the original PoseNet's architecture were also proposed, focusing on the encoder and localizer components (Table 1). Instead of using a GoogLeNet architecture, Hourglass-Pose [31] suggested an encoder-decoder (hourglass [32]) architecture implemented with a ResNet34 [33] encoder (removing the average pooling and softmax layers). In SVS-Pose [34], a VGG16 architecture [35] was used instead: convolutional layers replaced the GoogLeNet architecture (encoder) and its final first FC layer formed part of the localizer, together with 2 additional FC layers. Instead of using a shared localizer architecture, BranchNet [10] used a partial GoogLeNet architecture, truncated after the 5[th] Inception module (Icp), as a shared encoder and duplicated the remaining architecture (6[th]-9[th] Icps) and the FC localizer to form two separate localizer branches, for the translation and quaternion, respectively. Note that in all these architectural modifications, the regressor component remained unchanged (see Eq. 6, Fig. 1 and Section 1).

One interesting, and perhaps surprising, observation, made by the authors of PoseNet, was that learning the orientation and translation as separate tasks led to poor performance, compared to jointly learning them using a weighted loss (Eq. 6) [11]. However, the implication of this joint loss, is that the balancing factor $\beta$ needs to be set a-priori, requiring lengthy fine tuning (for example, though grid search) and resulting in different values for different conditions such as indoor and outdoor scenes [11]. To address this problem, Kendall and Cipolla suggested an alternative loss for optimizing PoseNet [11]. To better model the uncertainty associated with the pose estimation task, the researchers proposed a loss with learned uncertainty parameters (*learnable weights pose loss*):

$$\mathcal{L}_\sigma(I_c) = \mathcal{L}_x(I_c)\exp(-\hat{s}_x) + \hat{s}_x + \mathcal{L}_q(I_c)\exp(-\hat{s}_q) + \hat{s}_q \quad (7)$$

where $\hat{s}_x = \log(\hat{\sigma}_x^2)$ and $\hat{s}_q = \log(\hat{\sigma}_q^2)$ and $\hat{\sigma}_x^2$, $\hat{\sigma}_q^2$ are the variances (scalar values) implicitly learned from the losses of translation and rotation estimation tasks, respectively, through backpropagation [11]. Note that the choice of logarithm of the variance gives an improved stability by preventing a division by zero and constraining the variance values [11, 36]. The effect of using this loss formulation is a better (learned) balance between the different magnitude of the quaternion and translation vector and a regularization of the loss through the network's uncertainty. A second variant was also introduced as part of this work, using the reprojection error of the estimated pose. This *reprojecion error pose loss* is defined by projecting 3D points to 2D pixels with the ground-truth and estimated poses, and taking the mean of the Euclidian distances between the projected 2D points. Although this variant better captures the true error of the estimates pose, in practice, it led to instability and the learning did not converge. However, when used for fine-tuning a pre-trained PoseNet (first trained with the loss $\mathcal{L}_\sigma$), this loss led to an additional improvement [11].

Instead of using just the absolute pose error, Brahmbhatt *et al.* suggested to include additional data sources in order to constrain the loss [37]. A core network, named MapNet is first trained in a supervised manner with absolute and relative ground truth data. The loss in Eq. 7 is extended to include the relative pose loss:

$$\mathcal{L}_{\sigma_s}(I_c^1, I_c^2) = \mathcal{L}_\sigma(I_c^1) + \mathcal{L}_\sigma(I_c^1, I_c^2) \quad (8)$$

where $\mathcal{L}_\sigma(I_c^1, I_c^2)$ is defined over the relative losses $\mathcal{L}_x(I_c^1, I_c^2)$ and $\mathcal{L}_q(I_c^1, I_c^2)$, computed using the estimated and ground truth relative translation and quaternion (between $I_c^1$ and $I_c^2$) instead of the absolute ones. An extended version, MapNet+, make use of additional sensor measurements such as IMU and GPS and unlabeled video data by computing absolute and relative pose labels with appropriate algorithms (e.g., computing the relative loss between consecutive frames with visual odometry algorithms [37,38,39]). A pretrained MapNet is fine-tuned in a semi-supervised manner, on the labelled and unlabeled data, where the total loss is now a sum of the supervised loss (Eq. 8) and the loss for each data source $d$:

$$\mathcal{L}_{\sigma_{total}}(I_c^1, I_c^2) = \mathcal{L}_{\sigma_s}(I_c^1, I_c^2) + \sum_d \mathcal{L}_{\sigma_d}(I_c^1, I_c^2) \quad (9)$$

On inference, MapNet+'s absolute pose predictions can be further refined using visual odometry data with standard pose graph optimization (PGO) algorithms (this version is referred to as MapNet+PGA). It is worth noting that MapNet also introduced changes to the original PoseNet architecture (see Table 1). A modified ResNet34 (adding a global average pooling layer after the last conv layer) was used as an encoder, instead of GoogLeNet. The researchers have also reported that using the logarithm of a unit quaternion to represent the orientation, achieved better results, compared to the usual quaternion representation (for further details on the logarithm form of a quaternion see [40]).

Table 1 summarizes the different modifications to PoseNet's architecture, described above. For the benefit of the reader, we further provide a visualization of selected examples (Fig. 3).

| Pose Regressor | Encoder | Localizer |
|---|---|---|
| PoseNet [7] | GoogLeNet | 1 FC |
| LSTM-Pose [30] | GoogLeNet | 1 FC + 4 LSTM |
| Hourglass-Pose [31] | ResNet34 Encoder-Decoder | 1 FC |
| SVS-Pose [34] | VGG16 (conv layers) | 3 FC |
| BranchNet [10] | GoogLeNet (truncated after the 5$^{th}$ Icp) | 2 x [GooLeNet (6$^{th}$ - 9$^{th}$ Icp) + 1 FC] |
| MapNet [37] | ResNet34 + global average pooling | 1 FC |

**Table 1.** Pose regressors with different encoder and localizer architectures.

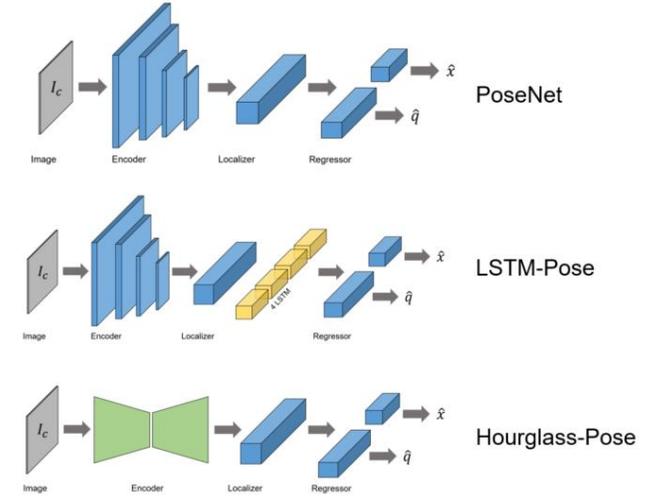

**Figure 3.** Example modifications to PoseNet's architecture.

*Auxiliary Learning*

Loss and architecture modifications to PoseNet's solution led to a significant improvement in its pose error for indoor and outdoor scenes (Table 3-4). However, even with this performance boost, pose regressors were not able to achieve comparable results to classic structure-based methods (see Tables 3-4).

Valada *et al.* attributed part of this subpar performance to the absence of 3D information from the learning process [41]. They proposed to learn additional auxiliary tasks, which share representations with absolute pose estimation, in order to maximize its learning. The corresponding network, named VLocNet, implemented an auxiliary learning approach, by jointly learning absolute pose estimation (the main task) with relative pose estimation (the auxiliary task). Given two images, $I_c^{t-1}, I_c^t$, captured at time $t-1$ and $t$, respectively, two ResNet50 [33] branches, truncated after the *conv3_x* layer (Res3 in VLocNet), are used to encode $I_c^{t-1}$ and $I_c^t$. Note that a shared branch is applied twice for $I_c^t$, and that an identical, yet separate branch is applied for $I_c^{t-1}$, forming a non-conventional 'Triamese' sub-network.

In order to learn the absolute pose of $I_c^t$ and directly leverage on visual odometry information, one of the encodings of $I_c^t$ (outputted by Res3 of the absolute pose subnetwork) and the previous pose $p_{t-1}$ (ground truth during training and estimated on test time) are passed through the *conv4_x* layer of ResNet50 (Res4 in VLocNet) and a FC layer, respectively. The output of the latter is a 200,704-dimensional vector, further reshaped, to match in dimensions the output of Res4. A modified *conv5_x* layer (Res5 in VLocNet, changed to take a 2048-dimensional-deep input) takes the concatenation (channel-wise) of these outputs. The resulting encoding is passed to a localizer and a regerssor components as in PoseNet. Note that this sub-network can be thought of as a pose regressor, whose encoder is a modified ResNet50, designed to include information from the previous pose.

The relative pose is regressed in a similar manner to the absolute pose. The second encoding of $I_c^t$ (output of Res3 in the relative pose subnetwork) and the encoding of $I_c^{t-1}$ are passed through two streams of a *conv4_x* layer (respectively) and concatenated (channel-wise). The concatenated output is passed to a modified *conv5_x* layer, and the resulting encoding is again fed to a localizer and regressor components to output the relative pose. For both the shared, absolute and relative sub-networks, the researchers have further modified the ResNet50 architecture by replacing RELU [42] layers with ELU [43] layers.

VLocNet was trained in an alternating manner, optimizing either the loss of the absolute pose or the loss of the relative pose (using two separate optimizers). The relative loss (*visual odometry loss*) is define as $\mathcal{L}_\sigma(I_c^{t-1}, I_c^t)$ (similarly to MapNet). The absolute loss (*geometric consistency loss*) is an extended version of the loss defined in Eq. (7), where $\mathcal{L}_x(I_c)$ and $\mathcal{L}_q(I_c)$ are extended (summation of losses) to include the translation and quaternion losses of the ground truth relative pose and the implied relative pose between the estimated absolute poses of $I_c^{t-1}$ and $I_c^t$ (the motion from the pose $p_{t-1}$ to $p_t$).

By harnessing the notion of auxiliary learning and introducing geometric constraints, VLocNet was able to achieve *state-of-the-art* results for indoor and outdoor scenes, reducing the error by an order of magnitude, compared to 'simple' pose regressors. Following this success, an extended version of VLocNet, named VLocNet++ [44], added semantic segmentation as a second auxiliary task. Specifically, the network architecture was extended to include a network dedicated to semantic segmentation, where intermediate outputs are exchanged with the absolute pose network. This addition led to a further improvement, surpassing classical structure-based methods on indoor scenes (see Table 3).

A related work, named DGRNets [45] built upon the VLocNet architecture but modified it to output only the absolute pose (of $I_c^t$). Specifically, the outputs of the modified *conv5_x* layers (in the absolute and relative networks) are fed to LSTM layers followed by a FC layer. The feature vectors from the FC layers of the absolute and relative sub-networks are then concatenated and passed as usual to a localizer and regressor components to regress the absolute pose. In addition, the geometric consistency loss was extended to a *temporal geometric consistency loss*, to include constraints from multiple pairs of rigid body transformations. The motivation of the authors was to leverage on LSTM units to capture temporal correlations (inspired by previous work on video [46]) and to directly embed knowledge from relative motion into the prediction of the absolute pose.

### 2.1.2 Hybrid Pose Learning

End-to-end deep learning offers a simple and appealing paradigm: a single architecture to learn a complex function mapping from input to output (in our case, an image and its camera's pose, respectively). However, learning an entire pipeline often results in subpar performance, when compared to learning each task in the pipeline separately [25,47]. In addition, in the context of pose estimation, end-to-end learning imposes a tight coupling to the scene coordinates, where the network can be viewed as a compressed version of an underlying map (of the scene) [48]. This in turn, limits the generalization power of the network. Hybrid pose learning methods shift the learning towards local or related problems, and combine them with the traditional image retrieval and structure-based pipelines.

*Image Retrieval with Relative Pose Regression*

Given a (query) anchor image $I_c^a$, image retrieval methods estimate its pose to be the pose of its nearest neighbor, $I_c^b$, in a reference database. The similarity metric defining the closeness of two images typically rely on visual similarity. Note that the ground truth pose $p_a = (x_a, q_a)$ of $I_c^a$ can be computed from the (ground truth) pose of $I_c^b$, $p_b = (x_b, q_b)$, and the relative pose from $I_c^a$ to $I_c^b$, $p_{a \to b} = (x_b - x_a, q_a^{-1} q_b)$, through simple arithmetic operations:

$$x_a = x_b - (x_b - x_a) \tag{10}$$

$$q_a = (q_a^{-1} q_b q_b^{-1})^{-1} \qquad (11)$$

Using the relationship defined in Eq. 10 and Eq. 11, researchers have proposed to learn the relative pose $p_{a \to b}$ and combine the estimate with the traditional image retrieval pipeline in order to predict the absolute pose $p_a$.

NNnet [49] used a Siamese ResNet34 architecture (truncated after the last average pooling layer) to encode the visual features of a pair of images. The resulting 512-dimensional feature vectors were concatenated to form a single feature vector passed to a (1024-dimensional) localizer and regressor components, similar to PoseNet, in order to estimate the relative pose. The resulting network was trained with a pose loss (Eq. 6) defined over the estimated and ground relative poses (*relative pose loss*).

ReLocNet [50], proposed a similar architecture, with a different loss, where feature vectors, encoded by the dCNN, are used both for regressing the pose (as in NNnet) and for computing a *camera frustum loss*. This loss is taken to be the norm of the difference between the true frustrum and estimated frustum (from the difference between the two feature vectors). The overall loss is then the summation of the camera frustum loss and the relative pose loss. In addition, ReLocNet used the rotation matrix representation, rather than the quaternion, for training and inference.

After training the network to regress the relative pose, one of the Siamese branches can be further used to encode a reference database. At inference time, a query image is encoded by the same branch and the nearest neighbor is fetched (image retrieval) using the dot product of the encodings as a similarity measure. The estimated pose of the query image can then computed from the ground truth pose of the neighbor and the estimated relative pose using Eq. 10 and 11.

While the above approach is described for the case of estimating the absolute pose from the first nearest neighbor, both NNnet and ReLocNet have further suggested methods for leveraging on information from $k$ nearest neighbors to improve the final estimate.

*Structure-based with local learning*

A structure-based pipeline for pose estimation employs RANSAC for refining a hypothesized pose. Given a (minimum of) $\lceil n/2 \rceil$ 2D-3D matches, a PnP solver is applied to generate a pose hypothesis. Given a set of pose hypotheses generated in this manner, each candidate pose is applied to the entire collection of 2D points and is assigned with a score to reflect the overall consistency (e.g., inlier count). The estimated pose is taken to be the pose with the maximal count. Brachmann *et al.* [51] suggested to perform the 2D-3D matching, traditionally done based on descriptor matching, by directly regressing the 3D coordinates from 2D pixels and couple this learning with a structure-based pipeline. While the idea of scene coordinates regression was not new (previously achieved with Random Forests [52]), it was still an independent pre-processing step before RANSAC was applied. Instead, the researchers formulated an alternative pipeline where one dCNN (VGG13 architecture) is trained to regress the 3D scene coordinates from (crops of) the 2D image pixels and a second dCNN (VGG13) is trained to select the best hypothesis. Since *argmax* and the inlier count operations are not differentiable, a Differentiable RANSAC (DSAC) was proposed, where a probabilistic selection is employed, according to a softmax distribution and the scoring function is learned. Similarly to traditional structure-based pipelines, the chosen pose hypothesis is further refined. The network was trained end-to-end, with both the *scene coordinates regression loss* (Euclidian distance between the true and estimated coordinates) and the pose loss. Here, instead of summing the losses of the translation and orientation, the pose loss was taken to be the maximal value between them. While offering a novel learning-based structure-based pipeline, DSAC's authors hypothesized that learning the scoring function hampers performance. In a following version, named DSAC++ [53], they suggested to learn only the scene coordinates regression task (while still utilizing the overall pose loss). As an alternative to learning the score, the sigmoid function was applied for computing a soft inlier count, $s(p)$, for a hypothesized pose $p$:

$$s(p) = \sum_i sig(\tau - \omega r_i(p)) \qquad (12)$$

where $r_i(p)$ is the reprojection error of point $i$ (when applying $p$), $\tau$ is the inlier threshold and $\omega$ controls the softness of the sigmoid. Similarly to DSAC, the score is used to sample from a softmax distribution. In addition, by introducing an initialization heuristics for the scene coordinates (assuming a depth prior and using the focal length) DSAC++ elevated the need for a 3D model. The overall optimization now consisted of three steps: (1) scene coordinates initialization, where a scene coordinate regression task is learned, with a scene coordinates regression loss (as in DSAC) (2) a further learning of the scene coordinate regression task, now minimizing the *reprojection error loss* (calculated with the ground truth pose) and (3) end-to-end optimization with the pose loss (as done in DSAC). The additional second step, enabled the network to implicitly learn the 3D geometry, without a 3D model. Other modifications to the initial cropping strategy and the final pose refinement procedure were also introduced as part of DSAC++ to improve DSAC's performance. Indeed, DSAC++ was able to surpass its counterpart feature-based structure-based methods for indoor and outdoor scenes (Tables 3-4).

*Hierarchical Pose Estimation*

Image retrieval methods are fast and simple but give a rough pose estimate. Structure-based methods require a 3D

model that grows linearly with the size of the scene. In addition, the search through the shared descriptor space (2D-3D matching), becomes slower and more prone to errors, as the scene grows (the search space grows and ambiguous matches are more likely). In a recent work [24], researchers have suggested a hierarchical paradigm to form a synergy between the two approaches:
1. Given an image $I_c$, use an image retrieval method to get its k-closest images (relying on global descriptors, computed offline), and identify small sub-scenes ('places'), by mapping the fetched images to connected components in the corresponding co-visibility graph (from SfM).
2. Given the candidate places, defining a smaller restricted search space, apply a structure-based pipeline to estimate the pose.

To leverage on advancements in computer vision and deep learning, this paradigm of coarse-to-fine localization can be implemented using *state-of-the-art* learned global (for image retrieval) and local (for 2D-3D matching) descriptors. As a proof-of-concept, the researchers used NetVLAD [21] for image retrieval and SuperPoint[19] for computing local descriptors. The resulting architecture achieved *state-of-the-art* results on several challenging datasets presenting variations in lighting, seasons and viewpoints [54]. In order to adapt the proposed paradigm to localization on mobile, a distilled alternative architecture, named HF-Net, was further introduced, achieving comparable (slightly reduced) results. Given an image $I_c$, a shared encoding is generated using a MobileNet [55] architecture. The shared features are then passed to two subnetworks, computing the global and local descriptors, respectively. The global descriptor is computed with a second MobileNet sub-network appended with a NetVLAD layer. A SuperPoint decoder is used instead to compute the local descriptors and keypoints scores. HF-Net was trained using multitask [36] distillation [56] with pretrained NetVLAD and SuperPoint as its teachers, where the loss is a weighted sum of the three losses (for three tasks):

- Squared Euclidian distance between the global descriptor predicted by the teacher NetVLAD ($d_s^g$) and the student HF-Net ($d_t^g$).
- Squared Euclidian distance between the local descriptor predicted by the teacher SuperPoint ($d_s^l$) and the student HF-Net ($d_t^l$).
- Cross entropy between the keypoint scores predicted by the teacher SuperPoint ($k_t$,) and the student HF-Net ($k_s$):

$$\mathcal{L}_H(I_c) = e^{-w1}\|d_s^g + d_t^g\|_2^2 + e^{-w2}\|d_s^l - d_t^{gl}\|_2^2 + 2e^{-w3}\text{CrossEntropy}(k_t, k_s) + \sum_i wi \quad (13)$$

where $w1, w2\ and\ w3$ are learnable parameters and $\sum_i wi$ is a regularization term.

While distillation addresses the need for a compact network, for running on limited memory devices, multi-task learning addresses the problem of limited training data, leveraging on shared representations to boost the performance on each task.

2.2. Results

In Section 2.1 we have described different approaches, highlighting representative methods. In order to further analyze the pros and cons of these approaches, we have complied cross-comparisons of reported pose errors and several algorithmic characteristics.

Table 2 lists 21 deep pose estimation methods, grouped by their algorithmic cluster (see Fig. 2) and several key properties: year of publication, loss function, inference time and a reference to the original implementation (if available). Tables 3 and 4 compare the pose error (Median) for these methods (if available), when tested against two datasets: 7Scenes [27] and Cambridge Landmarks [7] (correspondingly). We also include results for a *state-of-the-art* structure-based method, named ActiveSearch [9], as a reference. ActiveSearch implements a structure-based pipeline using SIFT as a descriptor, with an efficient search heuristics to reduce match time.

The 7Scenes dataset includes seven small-scale indoor scenes, with a spatial extent of a few squared meters. The Cambridge Landmarks dataset is a mid-scale urban outdoor dataset. It includes six scenes, ranging in their spatial extent from 35x25m (Shop Façade) to 500x100m (Street). We chose these two datasets as they are commonly used for evaluation of absolute pose estimators and as they encompass common challenges and scenarios typical of visual localization applications: mid/small-scale, indoor/outdoor, repeating elements, textureless features and significant viewpoint changes and trajectory variations between train and test sets.

When analyzing the results presented in Tables 3-4, a first observation is that pose regressors are inferior to other deep pose estimators. While incremental improvement (compared to PoseNet) is achieved through modifications to architecture and loss function (see also Section 1 – *Pose Regressors* and Table 1), a step-change in performance is achieved with other methods. In addition, even when coupled with relative pose regression, the image retrieval paradigm does not deliver a competing pose error either.

Instead, auxiliary learning (VLocNet, VLocNet++) and structure-based methods with local learning (DSAC, DSAC++) are able to achieve comparable results or surpass Active Search. HF-Net, which was not evaluated against the 7Scenes and Cambridge (and thus not included in Tables 3 and 4), was also able to surpass ActiveSearch (and DSAC+) on other datasets presenting various challenging conditions. As oppose to pose regressors, these methods rely on geometrical constraints, either by utilizing a structure-based pipeline (DSAC, DSAC++ and HF-Net) or by jointly learning both the absolute pose and the relative motion in the scene (VLocNet, VLocNet++). In the hybrid approach paradigm, the learning is focused on 'local' computer vision tasks (2D-3D matching in DSAC++, local and global descriptors in HF-Net). Auxiliary learning benefits from adding auxiliary tasks, which provide additional information about the scene and attend the network to important patches in the image (VLocNet, VLocNet++). In addition, employing a creative integration of tasks, losses and shared weights which capture different aspects about the scene, show better performance between variants of the same method. For example, by adding semantic segmentation as an additional auxiliary task, VLocNet++ is able to achieve a significant improvement, compared to its predecessor, VLocNet.

For all methods, pose error (translation and rotation) increases for outdoor scenes. Moreover, for very large scenes (Street scene in the Cambridge Landmarks dataset), leading methods such as VLocNet and DSAC++ failed to learn. One likely explanation is due to network capacity, limited by the spatial extent it can learn. In this context, HF-Net (successfully applied to large outdoor scenes) has an advantage thanks to its hierarchical approach. Although Hybrid approaches show *state-of-the-art* results (HF-Net and DSAC++), some with the ability to handle large-scale scenes (HF-Net), end-to-end deep learning methods have real-time inference times and an appealing simple pipeline (Table 2).

It is worth noting that an official implementation is available only for 66% of the reported methods. The availability of a code base significantly contributes to the reproducibility of results and to the ability of researchers to evaluate methods in a fair and consistent manner. In the next section we provide practical notes, based on our experience, to facilitate implementation and benchmarking.

| Method | Year | Approach | Loss | Inference time (ms) | Implementation |
|---|---|---|---|---|---|
| PoseNet [7] | 2015 | Pose Regressors | Pose Loss (Eq. 6) | 5 | Yes [7] |
| Dense PoseNet [7] | 2015 | Pose Regressors | Pose Loss (Eq. 6) | 95 | Yes [7] |
| Bayesian PoseNet [29] | 2016 | Pose Regressors | Pose Loss (Eq. 6) | 6 (128 models) | Yes [7] |
| LSTM-Pose [30] | 2017 | Pose Regressors | Pose Loss (Eq. 6) | NA | No |
| Hourglass-Pose [31] | 2017 | Pose Regressors | Pose Loss (Eq. 6) | NA | Yes [31] |
| SVS-Pose [34] | 2017 | Pose Regressors | Pose Loss (Eq. 6) | 12.5 | No |
| BranchNet [10] | 2017 | Pose Regressors | Pose Loss (Eq. 6) | 6 | No |
| PoseNet + Learnable weights pose loss [11] | 2017 | Pose Regressors | Learnable weights pose loss (Eq. 7) | 5 | Yes [7] |
| PoseNet + Reprojection error pose loss [11] | 2017 | Pose Regressors | Reprojection error pose loss | 5 | Yes [7] |
| MapNet [37] | 2018 | Pose Regressors | Eq. 8 | NA | Yes [37] |
| MapNet+ [37] | 2018 | Pose Regressors | Eq. 9 | NA | Yes [37] |
| MapNet+PGO [37] | 2018 | Pose Regressors | Eq. 9 + PGO optimization | NA | Yes [37] |
| VLocNet [41] | 2018 | Auxiliary Learning | Geometric consistency loss + visual odometry loss | NA | No |
| VLocNet++[44] | 2018 | Auxiliary Learning | Geometric consistency loss + visual odometry loss | 79 | No |
| DGRNets [45] | 2018 | Auxiliary Learning | Temporal geometric consistency loss | NA | No |
| NNnet [49] | 2017 | Image Retrieval w/ Relative Pose Regression | Relative pose loss | NA | Yes [49] |
| RelocNet [50] | 2018 | Image Retrieval with Relative Pose Regression | Relative pose loss + camera frustum loss | NA | No* |
| DSAC [51] | 2017 | Structure-based with local learning | Scene coordination regression loss + Pose Loss (max variant) | 1500 | Yes [51] |
| DSAC++ [53] | 2018 | Structure-based with local learning | Scene coordination loss + Reprojection Error pose loss + Pose Loss (max variant) | 200 | Yes [53] |
| NetVLAD+SuperPoint [54] | 2019 | Hierarchical | Pretrained | ~150 | Yes [54] |
| HFNet [54] | 2019 | Hierarchical | Eq. 13 | ~50 | Yes [54] |

**Table 2.** A listing of learning base pose estimators and some of their key properties. For each estimation method we list its name, year of publication, the approach it employs (mapped to one of the leaves in the hierarchical algorithmic segmentation, in Fig. 2), the loss or losses used for optimization (referring to the definitions in Section 2.1) and the inference time per image in ms. We also report whether an official implementation is available and if so, include a reference to the code base.
[*] link leads to an erroneous gateway

|  | **Chess** | **Fire** | **Heads** | **Office** | **Pumpkin** | **Kitchen** | **Stairs** |
|---|---|---|---|---|---|---|---|
| PoseNet | 0.32m, 8.12° | 0.47m, 14.4° | 0.29m, 12.0° | 0.48m, 7.68° | 0.47m, 8.42° | 0.59m, 8.64° | 0.47m, 13.8° |
| Dense PoseNet | 0.32m, 6.60° | 0.47m, 14.0° | 0.30m, 12.1° | 0.48m, 7.24° | 0.49m, 8.12° | 0.58m, 8.34° | 0.48m, 13.8° |
| Bayesian PoseNet | 0.37m, 7.24° | 0.43m, 13.7° | 0.31m, 12.0° | 0.48m, 8.04° | 0.61m, 7.08° | 0.58m, 7.54° | 0.48m, 13.1° |
| LSTM-Pose | 0.24m, 5.77° | 0.34m, 11.9° | 0.21m, 13.7° | 0.30m, 8.08° | 0.33m, 7.00° | 0.37m, 8.83° | 0.40m, 13.7° |
| Hourglass-Pose | 0.15m, 6.53° | 0.27m, 10.84° | 0.19m, 11.63° | 0.21m, 8.48° | 0.25m, 7.01° | 0.27m, 10.84° | 0.29m, 12.46° |
| BranchNet | 0.18m, 5.17° | 0.34m, 8.99° | 0.20m, 14.15° | 0.30m, 7.05° | 0.27m, 5.10° | 0.33m, 7.40° | 0.38m, 10.26° |
| PoseNet + Reprojection error pose loss | 0.13m, 4.48° | 0.27m, 11.3° | 0.17m, 13.0° | 0.19m, 5.55° | 0.26m, 4.75° | 0.23m, 5.35° | 0.35m, 12.4° |
| MapNet | 0.09m, 3.24° | 0.20m, 9.29° | 0.12m, 8.45° | 0.19m, 5.45° | 0.19m, 3.96° | 0.20m, 4.94° | 0.27m, 10.57° |
| VLocNet | 0.036m, 1.71° | 0.039m, 5.34° | 0.046m, 6.64° | 0.039m, 1.95° | 0.037m, 2.28° | 0.039m, 2.20° | 0.097m, 6.48° |
| VLocNet++ | 0.023m, 1.44° | 0.018m, 1.39° | 0.016m, 0.99° | 0.024m, 1.14° | 0.024m, 1.45° | 0.025m, 2.27° | 0.021m, 1.08° |
| DGRNets | 0.016m, 1.72° | 0.011m, 2.19° | 0.017m, 3.56° | 0.024m, 1.95° | 0.022m, 2.27° | 0.018m, 1.86° | 0.017m, 4.79° |
| NNnet | 0.13m, 6.46° | 0.26m, 12.72° | 0.14m, 12.34° | 0.21m, 7.35° | 0.24m, 6.35° | 0.24m, 8.03° | 0.27m, 11.80° |
| RelocNet | 0.12m, 4.14° | 0.26m, 10.4° | 0.14m, 10.5° | 0.18m, 5.32° | 0.26m, 4.17° | 0.23m, 5.08 | 0.28m, 7.53° |
| DSAC | 0.02m, 1.2° | 0.04m, 1.5° | 0.03m, 2.7° | 0.04m, 1.6° | 0.05m, 2.0° | 0.05m, 2.0° | 1.17m, 33.1° |
| DSAC++ | 0.02m, 0.5° | 0.02m, 0.9° | 0.01m, 0.8° | 0.03m, 0.7° | 0.04m, 1.1° | 0.04m, 1.1° | 0.09m, 2.6° |
| Active Search | 0.04m, 2.0° | 0.03m, 1.5° | 0.02m, 1.5° | 0.09m, 3.6° | 0.08m, 3.1° | 0.07m, 3.4° | 0.03m, 2.2° |

**Table 3.** Median translation (in meters) and rotation (in degrees) errors of different deep absolutes pose estimators, when tested on the 7Scenes dataset. The results for a *state-of-the-art* structure-based method (ActiveSearch) are shown as a reference.

|  | **Great Court** | **K. College** | **Old Hospital** | **Shop Façade** | **St M. Church** | **Street** |
|---|---|---|---|---|---|---|
| PoseNet | NA | 1.92m, 5.40° | 2.31m, 5.38° | 1.46m, 8.08° | 2.65m, 8.48° | 3.67m, 6.50° |
| Dense PoseNet | NA | 1.66m, 4.86° | 2.57m, 5.14° | 1.41m, 7.18° | 2.45m, 7.96° | 2.96m, 6.00° |
| Bayesian PoseNet | NA | 1.74m, 4.06° | 2.57m, 5.14° | 1.25m, 7.54° | 2.11m, 8.38° | 2.14m, 4.96° |
| LSTM-Pose | NA | 0.99m, 3.65° | 1.51m, 4.29° | 1.18m, 7.44° | 1.52m, 6.68° | NA |
| SVS-Pose | NA | 1.06m, 2.81° | 1.50m, 4.03° | 0.63m, 5.73° | 2.11m, 8.11° | NA |
| PoseNet + Reprojection error pose loss | 7.00m, 3.7° | 0.99m, 1.1° | 2.17m, 2.9° | 1.05m, 4.0° | 1.49m, 3.40° | 20.7m, 25.7° |
| VLocNet | NA | 0.836m, 1.42° | 1.07m, 2.411° | 0.593m, 3.53° | 0.631m. 3.91° | NA |
| DSAC | 2.80m, 1.5° | 0.30m, 0.5° | 0.33m, 0.6° | 0.09m, 0.40° | 0.55m, 1.6° | NA |
| LearnLess (DSAC++) | 0.4m, 0.2° | 0.18m, 0.3° | 0.20m, 0.3° | 0.06m, 0.30° | 0.13m, 0.4° | NA |
| Active Search | NA | 0.42m, 0.6° | 0.44m, 1.0° | 0.12m, 0.40° | 0.19m, 0.5° | 0.85m, 0.8° |

**Table 4.** Median translation (in meters) and rotation (in degrees) errors of different deep absolutes pose estimators, when tested on the Cambridge dataset. The results for a *state-of-the-art* structure-based method (ActiveSearch) are shown as a reference.

### 2.3. Practical Notes for Implementation and Deployment

When coming to test or implement an existing method (for example, in order to reproduce reported results, apply the method on a new dataset/use case or use it as a starting point for a new research direction) different practical issues often arise. Some of these issues might not be reported, or explained in details, in the literature as part of the method's description (due to space and effort considerations) and are left for the reader to figure out. In order to shorten the time to implementation, for researchers who are interested in pose estimation with deep learning, we have made an effort to describe architectures and losses in an explicit and detailed manner, even when these details were not given in the original description. In this section we further discuss a (non-exhaustive) list of general and method-specific issues as well as mitigations, when available (see also references to official implementations in Table 2).

Although not always explicitly mentioned, some pose repressors that use the quaternion representation for the camera orientation do not apply normalization during train time. Since there is no explicit constraint on the values, the estimated quaternion may not map to a valid rotation. As a result the orientation error (Eq. 8) may take a non-numeric value. However, in our experiments, we have found that when using a 'simple' pose loss (Eq. 6) and a PoseNet-like architecture, the norm of the estimated quaternions approaches to 1 over time. This is consistent with previous reported observations [7]. Nevertheless, it is worth noting that most methods (including an updated Torch implementation of PoseNet) do normalize the quaternion at train time. Independently of the latter, at test time, the estimated quaternions are normalized in order to avoid invalid orientation predictions (and evaluation).

Some methods require relative poses, in addition to the absolute ones, for their training. This requires additional pre-processing, where a sample is no longer an image and its pose but instead a pair of images, their poses and the relative pose between them. For example, given a scene of $N$ images (captured in a sequence), VLocNet and VLocNet++ require a sequential list of pairs $< I_c^{t-1}, I_c^t >$ for $t = 2, ..., N$. Note that images need to be sorted first by their capture time and that for scene with K sequences, we will have K (non-overlapping) sequential lists of pairs. In NNnet instead, pairs with enough visual overlap are used for training. In the original paper, the authors have measured the overlap by the percentage of pixels projected into the candidate image plane (using the respective ground truth pose and the depth maps)[1]. Alternatively, visual overlap can be quantified based on camera frustum information or by measuring the overlap between matches computed through Homograpy or projection (with the ground truth pose). In the context of relative pose information, the terminology referring to the relative motion is inconsistent and may sometimes be confusing. In all the architectures described in section 2.1, the estimated relative pose refers to the motion from a reference image to a target image. In a sequential scenario, this is the motion from $I_c^{t-1}$ to $I_c^t$ (see Eq. 10 and 11 for an explicit derivation).

Different learning-based pose estimators employ different losses and optimization strategies (see Section 2 and Table 2). The most 'simple' pose loss (Eq. 6) requires fine tuning of the β parameter. Several papers empirically found setting β to a value close to 1, yields advantageous results on indoor scenes. Kendall *et al.*, found that outdoor and indoor scenes are characterized by different β values, motivating a learnable weighted version of this loss (see Eq. 7). A different optimization strategy, used by VLocNet and VLocNet++, alternates between minimizing the relative and absolute poses using separate optimizers. When comparing this strategy to jointly optimizing the losses, VLocNet's authors found the alternation strategy yields better results [41].

In the next section we describe some of the open challenges for pose estimation with deep learning and conclude with a summary existing and emerging promising research directions.

---

[1] We thank the authors of NNnet for sharing this technical detail in an email correspondence.

3. Challenges and Future Directions

Absolute pose estimation involves general challenges that are not specific to deep learning (non-exhaustive):

- *Ground truth acquisition*: a basic requirement for evaluating any pose estimator is to have accurate ground truth poses. Acquiring such data typically involves expensive raw data collection (e.g., aligned rig of LiDar and cameras with accurate GPS) and software for reconstructing a 3D model (e.g. COLMAP, Bundler etc.) that is often time consuming, compute hungry and tedious to run.
- *Privacy*: accurate pose estimation typically requires a dense 3D model. This implies that such a model may give access to a detailed description of private scenes (e.g., bedroom). Furthermore, Pittaluga *et al.* have recently shown that even sparse 3D point cloud models from SfM preserve enough information to reconstruct the actual appearance of the scene [57]. In addition, in applications which infer the pose of an image captured by a user, it is reasonable to assume that individuals in the scene will be localized without their knowledge and/or permission.
- *Benchmarking*: while pose error is usually reported in the literature for cross-comparison of different pose estimators (see Table 2), additional key aspects, such as inference time and memory footprint, are not consistently published. This prevents comprehensive comparison for cases where such parameters are critical (e.g., real-time applications with limited resources).
- *Visual recognition*: scene dynamics as well as lightning, seasonal and view point changes can dramatically affect the visual appearance of a scene, making it hard to localize based on visual cues. In addition, textures and repeating elements present a challenge for methods which rely on local information for 2D-2D or 2D-3D matching.

Beyond the issues listed above, addressing pose estimation with deep learning introduces additional challenges, due to the specific nature of these methods. Several works have shown that deep pose estimators overfit the scene they were trained on [7, 25]. To address this problem (typical of many learning based methods), researchers have started to explore alternative loss functions that better capture the scene's geometrical and temporal constraints as well as to fuse other data sources (visual odometry, IMU, GPS) [37,41,44,45,50,51,53].

In addition to the visual recognition challenge (described above), differences between the camera used for acquiring the 3D model and the camera of the end-user may introduce variations between the distribution of the train and test datasets, leading to poor performance at inference time.

Utilizing neural style transfer and generative adversarial networks for data augmentation can enrich the scene training data in order to mitigate this problem [58,59].

Generalization to unseen scenes remains an open issue for current *state-of-the-art* deep pose estimators such as VLocNet++ and DSAC++. In order to bypass this issue, the entire target area should be included in the training process. To address capacity limitations of the network (see Section 2.2), the global scene can be broken to separate partially overlapping scenes that will be used to train separate models, which will be queried at test time (querying the relevant model, for example, based on rough localization).

With the extensive demand for deep learning on mobile devices, the inference time and memory footprint become an important factor. Following the success of HF-Net and recent advances in network distillation [60], we expect more real-time memory-light learning-based pose estimation methods to emerge. It is worth noting, that despite their importance, there is still a wide variation in how inference time and memory footprint are reported, if at all, in the description of deep pose estimation methods.

Common to all pose estimation methods described in this review is the use of transfer learning. However, several recent works have reported that in numerous cases, where a mid-size dataset of high quality is available, training from scratch yields advantageous results [61]. Whether this finding applies to deep pose estimation, remains an open question to date. Another common property, is the absence of a confidence score. As oppose to classification, where the output of a softmax function can be used as a probability surrogate, pose estimators do not provide a measure for their confidence. Recent advances in evaluating the uncertainty of a network [62] (in cases where softmax, or alike functions, are not in place) can be used to couple deep pose estimation with a localization confidence (and importantly indicate a likely failure to localize).

At the time of writing this review, three main approaches emerge as promising: (1) auxiliary learning (section 2.1.1, e.g., VLocNet++) (2) Hybrid hierarchical approaches (section 2.1.2, HF-Net) and (3) Structure-based methods with local learning (DSAC++). Although these methods take a significantly different approach to solving the pose estimation problem, common to all three of them is the use of geometrical information of the scene. This shared feature, suggests that the scene's geometry is key for accurate pose estimation. This is in agreement with a recent study [25], suggesting that pose regressors fall short behind structure-based methods since they do learn the geometry of the scene.

In order to address challenges in visual recognition, other information about the scene, beyond its geometry can be beneficial. For example temporal information, semantic segmentation and attention maps, can help in identifying salient features (or masking features/areas prone to ambiguity). In addition, auxiliary data, coming from IMU, GPS and other sensors can be harnessed to improve performance.

Although relatively new, the field of pose estimation with deep learning has rapidly evolved, thanks to general advances in deep learning and computer vision. We expect the evolution of end-to-end deep learning pose estimators (from PoseNet to VLocNet++) to proceed, pushing their performance forward and mitigating some of their limitations. Hybrid approaches, and in particular hierarchical approaches, can be extended to adopt advanced coarse-to-fine matching strategies.

The challenges discussed here call for creative and novel ideas to be investigated. We hope this review will provide interested readers with the necessary tools to join the research journey of camera pose estimation with deep learning.